\documentclass{article}

    \usepackage[preprint]{neurips_2025}

\usepackage[utf8]{inputenc} 
\usepackage[T1]{fontenc}    
\usepackage{hyperref}       
\usepackage{url}            
\usepackage{booktabs}       
\usepackage{amsfonts}       
\usepackage{amsthm}         
\usepackage{nicefrac}       
\usepackage{microtype}      
\usepackage{xcolor}         
\usepackage{microtype}
\usepackage{graphicx}
\usepackage{subfigure}
\usepackage{booktabs} 
\usepackage{wrapfig}

\usepackage{amsmath}
\usepackage{amssymb}
\usepackage{mathtools}

\usepackage{amsmath,amsthm}

\usepackage{algorithm}
\usepackage{float}

\usepackage[noend]{algpseudocode}

\usepackage{tikz}
\definecolor{circleborder}{HTML}{1E9771}
\definecolor{circlefill}{HTML}{bce8da}
\newcommand*\annot[1]{\tikz[baseline=(char.base)]{
\node[shape=circle,draw=circleborder,fill=circlefill,minimum size=13pt,inner sep=0pt] (char) {#1};}}
\definecolor{dt}{HTML}{54ab8b}

\definecolor{rf}{HTML}{e3591c}

\definecolor{gb}{HTML}{5b6ea4}

\definecolor{nn}{HTML}{b74891}

\definecolor{ours}{HTML}{3a68c4}
\newcommand{\ours}{\textcolor{ours}{\textsc{ShapShift}}}

\hypersetup{
    colorlinks,
    linkcolor={ours},
    citecolor={ours},
    urlcolor={ours}
}

\title{
\ours{}: Explaining Model Prediction Shifts\\with Subgroup Conditional Shapley Values
}

\author{%
Tom Bewley\quad Salim I. Amoukou\quad Emanuele Albini\\
\textbf{Saumitra Mishra}\quad \textbf{Manuela Veloso} \\
J.P. Morgan AI Research
}

\begin{document}

\maketitle

\vspace{-0.2cm}
\begin{abstract}
Changes in input distribution can induce shifts in the average predictions of machine learning models. Such prediction shifts may impact downstream
business outcomes (e.g. a bank's loan approval rate),
so understanding their causes can be crucial. We propose \ours{}: a Shapley value method for attributing prediction shifts to changes in the conditional probabilities of interpretable subgroups of data, where these subgroups are defined by the structure of decision trees. We initially apply this method to single decision trees, providing exact explanations based on conditional probability changes at split nodes. Next, we extend it to tree ensembles by selecting the most explanatory tree and accounting for residual effects. Finally, we propose a model-agnostic variant using surrogate trees grown with a novel objective function, allowing application to models like neural networks. While exact computation can be intensive, approximation techniques enable practical application. We show that \ours{} provides simple, faithful, and near-complete explanations of prediction shifts across model classes, aiding model monitoring in dynamic environments.
\end{abstract}

\vspace{-0.2cm}
\section{Introduction}

Machine learning models often encounter shifts in their input distributions, such as between training and test data, or over time in production environments. In turn, input shifts may cause shifts in model predictions, as measured by aggregate statistics such as the mean.
Prediction shifts in deployed models can influence downstream decisions (e.g. the rate at which a hospital gives diagnoses or a bank approves loans), and thus high-level business outcomes such as resource utilisation, productivity and profitability.
They can also signal important upstream changes in production environments that may warrant deeper investigation.
Therefore, for model developers, owners and auditors, it may be valuable to know not only \textit{that} a prediction shift has occurred, but \textit{why} it has occurred.

\begin{wrapfigure}{r}{8cm}
    \centering
    \includegraphics[height=1.25in]{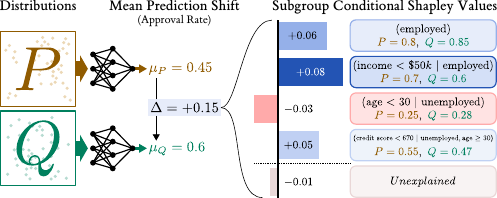}
    \vspace{-0.5cm}
    \caption{Using \ours{} to attribute
    an increase in the mean prediction of a loan approval model
    under a shift $P\rightarrow Q$
    to
    subgroup conditional probabilities. The reduction of working people earning $<\$50k$ has the largest positive impact. The small unexplained term quantifies the shift not accounted for by the four given factors. 
    }
    \label{fig:page1}
\end{wrapfigure}

\vspace{-0.1cm}
We propose to answer this \textit{why} question by attributing a given model's prediction shift to interpretable changes in its input distribution.
Our approach relies on the Shapley value (SV) framework \cite{shapley1953value}, which decomposes an observed prediction shift into the contributions of a set of factors (changes in the input distribution).
Specifically, we use \textit{subgroup conditional probabilities} as the elementary factors for which attributions are computed.
In the loan approval example, possible subgroup conditionals include the proportion of applicants in employment, $P(\text{employed})$, or the proportion of employed people who earn under $\$50k$, $P(\text{income}\ {<}\ \$50k\ |\ \text{employed})$.
Our method, called \ours{}, aims to identify a small, interpretable set of subgroup conditionals that, when used for attribution in the SV framework, explain the prediction shift of a given model between two distributions $P$ and $Q$ as fully as possible.
See Figure~\ref{fig:page1} for an illustrative example in the loan approval context.

\vspace{-0.1cm}
{
The core of \ours{} lies in leveraging the conditional structure of decision trees.
We first consider the simple case where the target model is itself a tree and show that any prediction shift can be fully explained by changes in subgroup conditional probabilities at the split nodes.
We then extend the method to tree ensembles by analysing individual trees and selecting the one whose structure best
\unskip\parfillskip 0pt \par
}

explains the ensemble's shift, accounting for residual effects.
This requires adding a novel `unexplained' term to the SV analysis.
Finally,
we propose an alternative, model-agnostic extension which attributes prediction shifts in arbitrary black box models via a surrogate tree.
This surrogate is grown with a novel objective function designed to directly
minimise the unexplained term, and thereby optimise the completeness of the resultant explanation.
While exact SV computation is feasible for small trees, its exponential complexity necessitates approximation strategies for larger models, which we also discuss.

\vspace{-0.1cm}
After presenting qualitative results on several
benchmark datasets, we perform a large-scale 
evaluation on hundreds of real-world distribution shifts in US Census data.
We find that \ours{} can rapidly compute simple, faithful and near-complete prediction shift explanations for both tree-based models and neural networks (via surrogates), balancing these desiderata more effectively than adapted baselines.
We see \ours{} as a significant contribution to the problem of model explanation in dynamic environments and a valuable addition to practical model monitoring pipelines.


\vspace{-0.1cm}
\section{Related Work}
\vspace{-0.1cm}

The problem of explaining the nature and effects of distribution shift is historically
under-explored \cite{babbar2024different}, but
is gaining
traction \citep{hinder2023one,klaise2020monitoring}.
Some methods aim to explain shifts in 
\textit{data}, while others explain the resultant effects on the predictions and performance of particular \textit{models}.

\vspace{-0.275cm}
\paragraph{Explaining Data Shifts}
Among diverse methods in this space, we highlight the work of Kulinski \& Inouye \cite{kulinski2023explaining}, which uses optimal transport (OT) to find a map between data from two distributions, followed by a clustering or sparsifying step to yield a more interpretable representation.
Also relevant is the work of Budhathoki et al. \cite{budhathoki2021why}, which uses SVs to attribute a difference between two marginal distributions of a target variable to conditional distributions of ancestor variables in a known causal graph.
We highlight points of connection to both of these works in later sections. 





\vspace{-0.275cm}
\paragraph{Explaining Prediction and Performance Shifts}
More directly related to our work are methods that explain the impact of distribution shifts on models.
For example, Shanbhag et al. \cite{shanbhag2021unified} use SVs
to attribute changes in models' predictive distributions
to groups of features and/or groups of inputs, which may be hand-designed using domain knowledge.
Later work by the same authors \cite{ghosh2022fair} uses similar techniques to explain changes in model fairness.
Taking a different perspective, Mougan et al. \cite{mougan2022explanation} analyse changes in the distribution of SHAP explanations \citep{lundberg2017unified} between datasets to detect model-impacting distribution shift and attribute it to particular features.

\vspace{-0.1cm}
Other works extend the aforementioned methods for explaining data shifts to analyse model performance changes.
Similar to Kulinski \& Inouye \cite{kulinski2023explaining}, Koebler et al. \cite{koebler2023towards} find an OT mapping between two distributions.
They then use it to estimate the resultant change in model performance and apply a SHAP-like analysis to obtain feature attributions.
Zhang et al. \cite{zhang2022why} adapt the causal graph intervention method of Budhathoki et al. \cite{budhathoki2021why} to compute SVs for conditionals with respect to a change in model performance,
and Feng et al. \cite{feng2024hierarchical} propose a similar method that only requires a partition of features into base and conditional subsets rather than a full causal model.
Still fewer assumptions are made by Cai et al. \cite{cai2023diagnosing}, who 
decompose
model performance degradation into shifts in inputs (covariate shift) and labels given inputs (concept shift).
Liu et al. \cite{liu2024need} build on this method by learning a decision tree to identify regions of input space with large concept shifts.
This work is one of several that find interpretable input regions where model-impacting shift occurs \cite{ackerman2021machine,ali2022lifecycle,d2022spotlight}, which bear resemblance to our subgroup representations defined by tree splits.

\vspace{-0.1cm}
Overall, prior work tackles many distinct problems, either relying on assumptions about causality \cite{zhang2022why},
shift structure \cite{koebler2023towards}
and domain knowledge \cite{shanbhag2021unified},
or giving only a coarse-grained decomposition of shifts and their effects \cite{cai2023diagnosing}.
We see a gap for a method explaining the specific phenomenon of prediction shift, providing granular insights
while making few domain-specific assumptions and explicitly quantifying the degree of explanatory completeness.

\section{Problem Statement: Attributing Prediction Shifts to Interpretable Factors}

\vspace{-0.1cm}
Let $f:\mathcal{X}\rightarrow\mathbb{R}$ be a model that maps inputs $x\in\mathcal{X}$ to real-valued predictions.\footnote{Our method can also be applied to more general models after a \textit{scalarisation} step; see Appendix~\ref{app:scalarisation}.}
Given an initial input distribution $P\in\Delta(\mathcal{X})$, the model has a mean prediction $\mu_P := \mathbb{E}_{x\sim P}[f(x)]$. When the distribution shifts to $Q\in\Delta(\mathcal{X})$, the mean prediction typically changes to $\mu_Q \neq \mu_P$.
We seek to explain the mean prediction shift $\mu_Q - \mu_P$ by attributing it to a set of \textit{factors} $C=\{c_1,...,c_n\}$, each of which is a measurable and human-interpretable property of a data distribution that differs between $P$ and $Q$.

\vspace{-0.1cm}
The SV framework provides a principled method for attribution, additively decomposing the prediction shift into a contribution $\phi_c$ from each factor, i.e. $\mu_Q - \mu_P=\sum_{c\in C}\phi_c$.
Applying the framework will require us to construct \textit{interventional} distributions by selectively swapping factors from their states under $P$ to their states under $Q$.
Formally, given a factor subset $S\subseteq C$, the associated interventional distribution should take the state of each $c\in S$ from $Q$ and take the remainder (for each $c'\in C\setminus S$) from $P$.
We will also require a means of evaluating the mean model prediction under this intervention, denoted by $\hat{\mu}_S$.
We can then compute each $\phi_c$ via the standard SV formula, which averages factor $c$'s contribution to the mean prediction over all interventions on the other factors:
\begin{equation}
    \label{eq:shapley}
    \phi_{c} \coloneqq \sum_{S\subseteq C\setminus\{c\}}\frac{|S|!(|C|-|S|-1)!}{|C|!}\Big[\hat{\mu}_{S\cup\{c\}} - \hat{\mu}_{S}\Big].
\end{equation}
A key contribution of this work is to define a factor set that is amenable to such an SV analysis.

\vspace{-0.1cm}
One intuitive approach might be to model the $P\rightarrow Q$ shift in terms of changes in the probabilities of data subgroups $g\subset \mathcal{X}$.
Returning to the loan model example, if $g_1$ represents the subgroup ``$\text{employed}=True$'' and $g_2$ represents the subgroup ``income < \$50k'', one might seek to quantify how a change in the joint probability $P(g_1, g_2)\rightarrow Q(g_1, g_2)$ has impacted the mean prediction. However, using joint probabilities directly as factors in an SV analysis presents significant challenges. As mentioned above, the SV calculation requires simulating the effect of changing one factor from its $P$-state to its $Q$-state while others remain fixed. If we change a single joint probability, the sum of all disjoint joint probabilities must still equal $1$. This necessitates an arbitrary, and often problematic, renormalisation of other joint probabilities, leading to ill-posed SVs where the attributions depend on the chosen renormalisation scheme.
Furthermore, changes in joint probabilities can be hard to interpret: do $P(g_1, g_2)$ and $Q(g_1, g_2)$ differ because $g_1$ became more prevalent overall, or because $g_2$ became more common \textit{within} $g_1$?
To overcome these issues, we propose to use changes in subgroup \textit{conditional} probabilities as the fundamental factors for attribution.
Instead of $P(g_1, g_2)$, we would factorise this into, for instance, $P(g_1)$ and $P(g_2|g_1)$.
The factors in our SV analysis are then the observed shifts in these conditional probabilities when moving from $P$ to $Q$.
We discuss more reasons for favouring an analysis of subgroup conditionals in Appendix \ref{app:why_subgroup}.

\vspace{-0.1cm}
Therefore, \ours{} explanations consist of (1) a carefully selected set of subgroup conditional probabilities whose values differ between $P$ and $Q$, and (2) the SVs attributing $\mu_Q - \mu_P$ to the change in each of these conditional probabilities. The practical challenge then becomes identifying a small, interpretable, yet sufficiently comprehensive set of such subgroup conditionals.
We aim to show that the structure of decision trees provides a natural and effective way to solve this selection problem.
To do so, we first consider a simple special case: where the target model $f$ is itself a decision tree.

\vspace{-0.2cm}
\section{
\ours{}
for Decision Tree Models}

\vspace{-0.1cm}
A decision tree is a hierarchy of split nodes, terminating in leaves that each specify a prediction. The tree's prediction $f(x)$ for an input $x\in\mathcal{X}$ depends on the leaf it reaches, which in turn depends on binary tests applied at the ancestor split nodes.
Consequently, the tree's mean prediction $\mu_P$ under some distribution $P$ is determined by the probability that the test at each split node evaluates to \textit{True}, given that this node is reached.
In the language of subgroups, these probabilities represent the proportion of data in a subgroup $g_2\subset\mathcal{X}$ defined by each split node,
conditional on belonging to another subgroup $g_1\subseteq\mathcal{X}$ defined by its ancestors.
Any change in the mean prediction under a distribution shift $P\rightarrow Q$ is entirely due to changes in these split node conditionals (e.g. $P(g_2|g_1)\rightarrow Q(g_2|g_1)$), and we aim to quantify their individual contributions using the SV framework.

\begin{figure*}
    \centering
    \includegraphics[width=\textwidth]{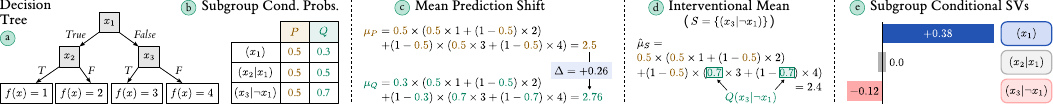}
    \vspace{-0.5cm}
    \caption{Calculation of SVs for subgroup conditionals to explain prediction shift in a decision tree.}
    \label{fig:single_tree_illustrative}
\end{figure*}

\vspace{-0.1cm}
Consider the toy example in Figure~\ref{fig:single_tree_illustrative}.
Subplot \smash{\annot{a}} shows a decision tree with $4$ leaves and $3$ split nodes over binary input features $x_1$, $x_2$ and $x_3$, \smash{\annot{b}} shows the associated split node conditional probabilities for two hypothetical distributions $P$ and $Q$, and~\smash{\annot{c}} derives the mean prediction shift $\mu_Q - \mu_P = +0.26$.
Our objective is to decompose this prediction shift into an SV attribution for each of the three split node conditionals, which serve as the factor set $C = \{(x_1), (x_2|x_1), (x_3|\neg x_1)\}$.

\vspace{-0.1cm}
To do so, we construct interventional distributions that mix the conditional probabilities from $P$ and $Q$, and compute what the tree's mean prediction would be under these interventions.
Concretely, for any subset $S\subseteq C$, the associated interventional distribution takes the conditional probability for each $c\in S$ from $Q$ and takes the remainder from $P$.
Simple multiplication allows us to derive the proportion of data
that would fall
in each leaf under this distribution, and thereby obtain the tree's mean prediction $\hat{\mu}_S$.
See Figure~\ref{fig:single_tree_illustrative}~\smash{\annot{d}} for an example with $S=\{(x_3|\neg x_1)\}$.
To compute an SV $\phi_c$ for each $c\in C$, we must perform \textit{all} such interventions and average over them via Equation~\ref{eq:shapley}.
The resultant SVs can be interpreted as the marginal impact of the change in each conditional probability from $P$ to $Q$ on the tree's mean prediction, averaged over all subsets of changes in the other probabilities.
We give pseudocode for the full SV calculation in Appendix~\ref{app:pseudocode}.

\vspace{-0.1cm}
Figure~\ref{fig:single_tree_illustrative}~\smash{\annot{e}} shows results for the toy example.
The positive prediction shift of $+0.26$ is entirely driven by the decreased probability that $x_1$ is \textit{True}, with the increased probability of $x_3$ given $\neg x_1$ having a contravening effect.
Since $P(x_2|x_1)=Q(x_2|x_1)$, the SV for this conditional must be zero.

\vspace{-0.1cm}
This basic approach resembles prior work that explains distribution shifts by swapping conditional probabilities in a causal graph \cite{budhathoki2021why,zhang2022why} and indeed, a decision tree can be viewed as a kind of causal graph for its own predictions (see Appendix~\ref{app:causal_graph}).
As with any SV method, the complexity is exponential in the number of
factors
(i.e. the number of split nodes), but once the conditional probabilities for both $P$ and $Q$ have been pre-computed from data, there is no further scaling with dataset size or input space dimensionality since each $\hat{\mu}_S$ is a closed-form function of these probabilities.
We discuss sampling and pruning methods for enhancing computational efficiency later.

\vspace{-0.1cm}
To demonstrate the method's applicability to real datasets, we consider a $12$-leaf tree for the \textit{California House Prices} dataset, shown in Figure~\ref{fig:single_tree_experiment}~\smash{\annot{a}} (see Appendix~\ref{app:expt_illustrative} for experiment details).
We define distribution $P$ as houses in low-income areas, and $Q$ as those in high-income areas.
The associated subgroup conditional probabilities are shown in Figure~\ref{fig:single_tree_experiment}~\smash{\annot{b}}.
Shifting from $P$ to $Q$ induces a prediction shift of $\mu_Q-\mu_P=+\$60.4k$.
Applying our method (Figure~\ref{fig:single_tree_experiment}~\smash{\annot{c}}), we find that the SVs for two conditionals are larger than the rest.
Over half of the prediction shift is attributed to an increase in the proportion of houses with many (more than $6$) rooms and, more subtly, a \textit{decrease} in the proportion of these many-roomed houses that have more than one bedroom.
The latter factor may indicate more rooms dedicated to leisure activities, which the model has learnt to associate with expensive properties.
Some other conditionals pertain to differences in the geographic
distribution of low- and high-income areas.
For example, the third highest SV is for the following conditional: among areas of small, high-occupancy houses south-east of San Francisco, more of the high-income ones lie in the South around Los Angeles, where predicted prices are higher (see Figure~\ref{fig:single_tree_experiment}~\smash{\annot{d}}).

\begin{figure*}[h!]
    \centering
    \includegraphics[width=\textwidth]
    {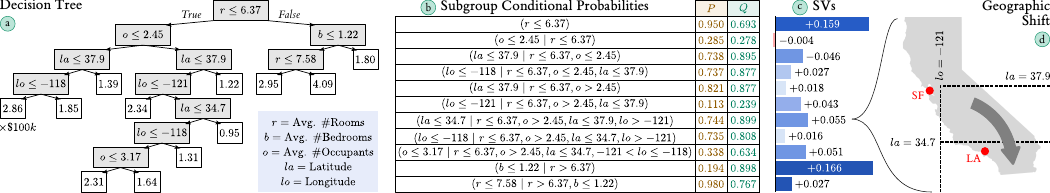}
    \vspace{-0.5cm}
    \caption{Application of \ours{} to a decision tree for house price prediction.}
    \label{fig:single_tree_experiment}
\end{figure*}


\vspace{-0.3cm}
\section{Extension \#1: \ours{} for Tree Ensembles}


\vspace{-0.1cm}
Tree ensembles, such as random forests \cite{breiman2001random} and gradient boosted trees \cite{friedman2001greedy}, are far more commonly used than single decision trees.
We now describe how our method can be extended to such models.

\vspace{-0.1cm}
Firstly, consider what would change if the $4$-leaf tree from Figure~\ref{fig:single_tree_illustrative} were part of an ensemble $f$.
The ensemble's mean prediction $\mu_P$ on a distribution $P$ still partly depends on the conditional probabilities at this tree's split nodes, but this information no longer \textit{fully} determines the mean prediction as we also need to know how the data are distributed in the \textit{other} trees.
This might seem like a major problem for our approach, necessitating an intractable joint analysis of the entire ensemble, However, we identify a small modification that enables us to still compute valid SVs at the single-tree level, and use them to explain ensemble-level prediction shifts, allowing for a degree of incompleteness.

\begin{figure*}
    \centering
    \includegraphics[width=\textwidth]{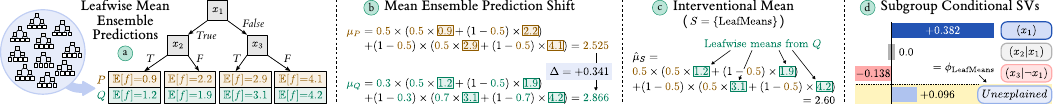}
    \vspace{-0.5cm}
    \caption{SV analysis of one member of a tree ensemble to explain ensemble-wide prediction shift, including an unexplained term to quantify the degree of incompleteness.}
    \vspace{-0.2cm}
    \label{fig:ensemble_illustrative}
\end{figure*}

\vspace{-0.1cm}
Our insight is that a shift in the mean prediction of an ensemble can be expressed using the conditional probabilities for any single tree, provided we also
account for the mean ensemble prediction in each leaf $l$ of that tree potentially differing between $P$ and $Q$, i.e. $\mathbb{E}_{x\sim P}[f(x)|x\in l]\neq\mathbb{E}_{x\sim Q}[f(x)|x\in l]$.
Such differences are due to changes in conditional probabilities across the rest of the ensemble, but the details of these changes can be abstracted away in this single-tree analysis.
Figure~\ref{fig:ensemble_illustrative}~\smash{\annot{a}} gives illustrative values of these differing leafwise means for a $4$-leaf ensemble member.
In turn, Figure~\ref{fig:ensemble_illustrative}~\smash{\annot{b}} expresses the ensemble's prediction shift of $+0.341$ in terms of this tree's conditional probabilities and leafwise means, assuming the same $P$ and $Q$ distributions as in Figure~\ref{fig:single_tree_illustrative}.

\vspace{-0.1cm}
To attribute this prediction shift to the subgroup conditionals of the single tree, we must incorporate the change in leafwise means between $P$ and $Q$ as an \textit{extra factor} in the SV analysis.\footnote{
One could also define a separate factor for each leaf, but this would be more costly; see Appendix~\ref{app:per_leafwise_mean}.
}
That is, we now compute the interventional mean $\hat{\mu}_S$ for each subset of interventions $S \subseteq (C\cup \{\text{LeafMeans}\})$, where the new $\text{LeafMeans}$ intervention swaps all leafwise means from their values under $P$ to their values under $Q$, as shown in Figure~\ref{fig:ensemble_illustrative}~\smash{\annot{c}}.
We then apply Equation~\ref{eq:shapley} to compute an SV $\phi_c$ for each $c\in C$, alongside an extra SV $\phi_\text{LeafMeans}$.
We refer to this as the \textit{unexplained} term because it quantifies the residual prediction shift left unaccounted for by the conditionals of this single tree.

\vspace{-0.1cm}
Results for the example are in Figure~\ref{fig:ensemble_illustrative}~\smash{\annot{d}}.
As in the single-tree case, the decreased probability of $x_1$ contributes positively to the prediction shift and the increased probability of $x_3$ given $\neg x_1$ contributes negatively.
However, these factors no longer provide a complete explanation because the unexplained term $\phi_\text{LeafMeans}$ is nonzero. We can quantify the degree of incompleteness with the following metric:
\vspace{-0.1cm}
\begin{equation}
    \text{PercentUnexplained} \coloneqq 100\times |\phi_\text{LeafMeans}|\ /\ |\mu_Q - \mu_P|,
\end{equation}
whose name is inspired by a related metric used in \cite{kulinski2023explaining}.
In the example, $\text{PercentUnexplained}=100\times 0.096/0.341=28.3\%$, indicating that changes in this tree's conditional probabilities account for just under $3/4$ of the total ensemble prediction shift.
This number is promising, but suggests room for improvement.
Indeed, such improvement is the aim of the second part of our extended method: rather than picking an arbitrary single tree from the ensemble on which to perform the SV analysis, we perform it on \textit{every} tree individually, and select the one with the smallest $\text{PercentUnexplained}$.
Doing so maximises the completeness of the resultant explanation.

\begin{figure*}[h!]
    \centering
    \vspace{-0.1cm}
    \includegraphics[width=\textwidth]{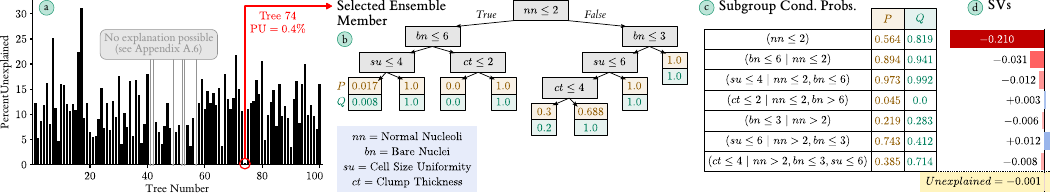}
    \vspace{-0.5cm}
    \caption{Application of \ours{} to a random forest for breast cancer malignancy prediction.}
    \label{fig:ensemble_experiment}
\end{figure*}

\vspace{-0.1cm}
Figure~\ref{fig:ensemble_experiment} illustrates this selection process on a random forest of $100$ trees, each with $8$ leaves, for the \textit{Breast Cancer Wisconsin}
dataset (see Appendix~\ref{app:expt_illustrative} for details).
We define distribution $P$ as all cases reported in January 1989, and $Q$ as those from later dates.
Shifting from $P$ to $Q$ induces a prediction shift of
$-0.253$
i.e. a $25.3\%$ reduction in the predicted malignancy rate.
Figure~\ref{fig:ensemble_experiment}~\smash{\annot{a}} shows the $\text{PercentUnexplained}$ for each member of the ensemble, which ranges from $31.2\%$ for tree $17$ to just $0.4\%$ for tree $74$. We therefore select tree $74$ as the most complete explanation.
Figure~\ref{fig:ensemble_experiment}~\smash{\annot{b}} and \smash{\annot{c}} show the structure of this tree, as well as its leafwise means and subgroup conditional probabilities

under $P$ and $Q$.
The plot of SV results in \smash{\annot{d}} reveals that the prediction shift can be largely explained by a \textit{single} factor: an increase in the proportion of cases with $\leq 2$ normal nucleoli.

\vspace{-0.1cm}
Since this extended method involves analysing trees independently, its complexity is only linear in the ensemble size, meaning the experiment in Figure~\ref{fig:ensemble_experiment} ran in just $1.13$ seconds on a \texttt{t3.2xlarge} AWS instance.
In Appendix~\ref{app:extra_gb}, we report equivalent results for a gradient boosted tree ensemble, for which we posit that an early stopping mechanism may be used to lower the computational expense.

\vspace{-0.2cm}
\section{Extension \#2: Model-Agnostic \ours{} via Surrogate Models}
\label{sec:model_agnostic}

\vspace{-0.1cm}
We have shown how white box access to a tree-based model $f$ allows us to explain its prediction shifts via subgroup conditionals.
We now consider the case where $f$ is another class of model, e.g. a neural network.
Following a common theme in model-agnostic interpretability \citep{molnar2020interpretable}, our basic strategy is to learn a decision tree \textit{surrogate}, to which we apply the same SV analysis as for a single member of a tree
ensemble. That is, we measure the mean prediction of $f$ in each leaf of the surrogate under $P$ and $Q$, and supplement the subgroup conditional SVs with an unexplained term capturing the effect of changes in these leafwise means.
As in the tree ensemble case, minimising
the $\text{PercentUnexplained}$
implies a more complete explanation.
We can achieve this by controlling how the surrogate is learnt.

\vspace{-0.1cm}
Decision tree learning follows a recursive growth process, whereby each extant leaf $l$ is split into two child leaves $l_\text{left}\cup \l_\text{right}=l$ with the aim of minimising their total \textit{impurity} $I(l_\text{left}) + I(l_\text{right})$.
Surrogate trees are typically grown with the same leaf impurity functions as for regular supervised learning, such as the variance in the labels of contained data for regression or the Gini impurity for classification.
The only difference is that the `labels' are defined as the predictions of the target model $f$.
To concretise this approach in our context, let $\mathcal{P}=\{x_i\sim P\}_{i=1}^N$ and $\mathcal{Q}=\{x_i\sim Q\}_{i=1}^M$ be datasets sampled i.i.d. from $P$ and $Q$ respectively, $\mathcal{D}=\mathcal{P} \cup \mathcal{Q}$ be the concatenation of the two, and $\mathcal{D}_l=\{x\in \mathcal{D}: x\in l\}$ denote the data falling in leaf $l$. Growing a surrogate tree with Gini impurity amounts to using $I(l)=\sum_{x\in\mathcal{D}_l}\sum_{x'\in\mathcal{D}_l}[f(x)\neq f(x')]$.
Crucially, however, we find that this approach creates suboptimal trees for our task of prediction shift explanation, often yielding unusably high $\text{PercentUnexplained}$ values. This is because the objective of na\"ive surrogate tree growth is not directly aligned with that of minimising the $\text{PercentUnexplained}$.

\begin{figure*}[h!]
    \centering
    \includegraphics[width=\textwidth]{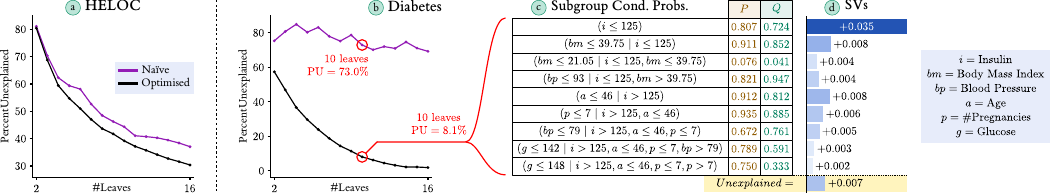}
    \vspace{-0.5cm}
    \caption{Comparison of na\"ive and optimised split criteria for surrogate growth.}
    \vspace{-0.15cm}
    \label{fig:model_agnostic_experiment}
\end{figure*}

\vspace{-0.1cm}
Instead, we propose a novel impurity function that provides a proxy for $\text{PercentUnexplained}$ minimisation, without requiring the costly computation of SVs in an inner loop during surrogate growth.
To motivate this proposal, we observe that the $\text{PercentUnexplained}$ is large when $|\phi_\text{LeafMeans}|$ is large, which in turn occurs when the mean prediction of $f$ in one or more leaves of the surrogate differs significantly between $P$ and $Q$.
Furthermore, the impact of each leafwise mean difference on the SV calculation will be large if the proportion of data in that leaf under distribution $P$ (denoted by $P(l)$) is large, \textit{or} the proportion under $Q$ (denoted by $Q(l)$) is large.
Therefore, we can expect $\text{PercentUnexplained}$ to be positively correlated with the following proxy quantity:
\begin{equation}
    \textstyle\sum_{l}\Big(P(l) + Q(l)\Big) \times \Big|\mathbb{E}_{x\sim P}[f(x)|x\in l] - \mathbb{E}_{x\sim Q}[f(x)|x\in l]\Big|.
\end{equation}
A more thorough analysis of this proxy, and its relationship to the SV calculation, is given in Appendix~\ref{app:impurity_as_proxy}.
In practice, we directly minimise the proxy by growing the surrogate with the following impurity function given per-distribution datasets $\mathcal{P}=\{x_i\sim P\}_{i=1}^N$ and $\mathcal{Q}=\{x_i\sim Q\}_{i=1}^M$:
\begin{equation}
    I(l) \coloneqq \left(\frac{|\mathcal{P}_l|}{|\mathcal{P}|} + \frac{|\mathcal{Q}_l|}{|\mathcal{Q}|}\right) \times \left|\frac{\sum_{x\in\mathcal{P}_l} f(x)}{|\mathcal{P}_l|}-\frac{\sum_{x\in\mathcal{Q}_l} f(x)}{|\mathcal{Q}_l|}\right|.
\end{equation}
To compare this optimised surrogate growth method to the na\"ive approach using Gini impurity, we combine both with our SV analysis to explain prediction shift in neural networks on the \textit{Home Equity Line of Credit} (HELOC) and \textit{Pima Indians Diabetes} datasets (see Appendix~\ref{app:expt_illustrative} for details).
In Figure~\ref{fig:model_agnostic_experiment} \smash{\annot{a}} and \smash{\annot{b}}, we plot the number of leaves in each surrogate against the $\text{PercentUnexplained}$ of the resultant SV explanation.
The optimised method brings an advantage across all tree sizes: moderate for HELOC, but very large for Diabetes.
Focusing on surrogates with $10$ leaves (hence $9$ conditionals) for Diabetes, the $\text{PercentUnexplained}$ is
$73\%$ for the na\"ive growth method, but just $8.1\%$ for the optimised method.
In this experiment, $P$ is defined as individuals with a low diabetes risk based on family history and $Q$ as those with a high risk. The neural network's prediction shift is $\mu_Q-\mu_P=+0.083$.
The subgroup conditionals and associated SVs for the optimised $10$-leaf surrogate, shown in Figure \ref{fig:model_agnostic_experiment} \smash{\annot{c}} and \smash{\annot{d}}, attribute the majority of this shift to a decrease in the rate of insulin measurements $\leq 125$, a decrease in the proportion of individuals with BMI $\leq 39.75$ below this insulin threshold, and a decrease in the proportion aged $\leq 46$ over the threshold.

\section{Quantitative Evaluation}

\vspace{-0.1cm}
Having shown qualitative results of \ours{} on benchmark datasets, we now perform a large-scale evaluation on real-world distribution shifts in five datasets from the \textit{Folktables} repository \citep{ding2021retiring}.
These datasets concern the prediction of an individual's employment, income, health insurance coverage, residential mobility or commute time using information from the US Census.
Crucially, there are data for all US states
(plus Puerto Rico) across five years from $2014$ to $2018$,
allowing us to study shifts across both geography and time.
All results in this section are aggregated over shifts between $50$ random pairs of (state, year) combinations for each dataset, for a total of $250$ shifts.
We highlight some key results in \textbf{bold} below, and give full experimental details in Appendix~\ref{app:expt_folktables}.




\vspace{-0.275cm}
\paragraph{Main Results}

We use \ours{} to explain \textit{Folktables} prediction shifts in decision trees with $8$ leaves (DT), random forest (RF) and gradient boosted (GB) ensembles of $100$ trees with $8$ leaves, and neural networks with a hidden layer of $100$ neurons (NN), for which we learn $8$-leaf decision tree surrogates.
Hence, all explanations are derived from $8$-leaf trees, comprising $7$ conditional factors.

\begin{figure*}[h!]
    \centering
    \includegraphics[width=\textwidth]{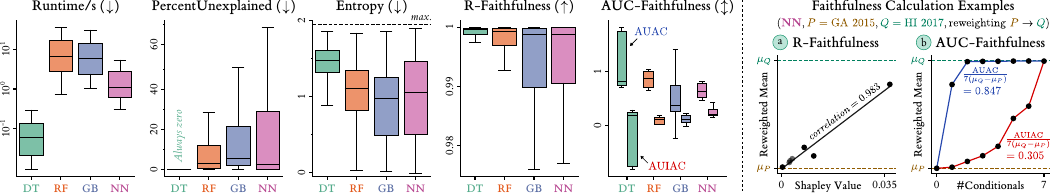}
    \vspace{-0.5cm}
    \caption{Aggregated performance metrics of \ours{} on $250$ shifts across five \textit{Folktables} datasets. $\downarrow$ / $\uparrow$ / $\updownarrow$ = lower / higher / more separated is better.}
    \label{fig:headline_results}
    \vspace{-0.25cm}
\end{figure*}

\vspace{-0.1cm}
The first two plots of Figure \ref{fig:headline_results} show distributions of runtimes
and $\text{PercentUnexplained}$ values for all model classes.
Runtimes of the model-agnostic variant (applied to NNs) are less than the tree ensemble variant, which is linear in the ensemble size.
\textbf{$\text{PercentUnexplained}$ averages below $6\%$ for \textit{all} models} (always zero for DTs), which is highly encouraging.
The upper ends of the distributions increase from left to right, indicating that near-complete explanations are occasionally unattainable.

\vspace{-0.1cm}
The next two plots report metrics from prior work \cite{bhatt2021evaluating,koebler2023towards}.
The first is the entropy of the absolute SVs $\{|\phi_c|:c\in C\}$, which has a maximum of $\ln(7)$ for $7$ factors.
Lower entropy implies a more parsimonious explanation, with more SV weight on fewer factors.
The distributions for RFs, GBs and NNs are similar, averaging around half the maximum and sometimes getting close to zero (indicating just one dominant SV).
Interestingly, the entropy of DT explanations is higher, potentially because they must always be complete.
The second metric from the prior work, which we call \textit{R-Faithfulness}, is the Pearson correlation ($R$) between each subgroup conditional SV and the corresponding effect on the model's mean prediction of reweighting data from $P$ to match the probability from $Q$ or vice versa. See Appendix~\ref{app:faithfulness_metrics} for details on this reweighting, and Figure \ref{fig:headline_results} \smash{\annot{a}} for an example of the $R$ calculation.
Results on this metric are very strong (note y-axis scale), with medians above $0.998$ for all models, and only occasionally dropping below $0.98$ for GBs and NNs.

\vspace{-0.1cm}
To verify that these results are not an artefact of the R-Faithfulness metric, we also perform an activation curve (AC) analysis \citep{schnake2021higher}, alongside the inverse variant (IAC) proposed by Kariyappa et al. \cite{kariyappa2024progressive}.
This involves progressively reweighting data to match each conditional in decreasing (AC) or increasing (IAC) order of SVs, and measuring the AUCs of the cumulative changes in mean model prediction (see Figure \ref{fig:headline_results} \smash{\annot{b}}).
For correctly-ordered attributions, the AUAC will exceed the AUIAC.
We plot the distributions of AUAC and AUIAC and find them to be almost fully separated for all model classes,
with significance below $p=10^{-50}$ under a one-sided Mann-Whitney U test.
This reinforces that \textbf{\ours{} yields faithful explanations, whose SV magnitudes reliably track the causal impact of subgroup conditional probabilities on mean model predictions}.

\begin{wrapfigure}{r}{2.649cm}
    \centering
    \includegraphics[width=2.649cm]{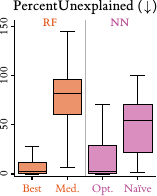}
    \vspace{-0.5cm}
    \caption{Impact of tree choice on $\text{PercentUnexplained}$.
    }
    \vspace{-0.3cm}
    \label{fig:tree_choice}
\end{wrapfigure}

\paragraph{Impact of Tree Choice}

In Figure \ref{fig:tree_choice}, we study the importance of two key design decisions in the ensemble-focused and model-agnostic variants of \ours{} (applied to all $250$ shifts).
If we select the tree with the median $\text{PercentUnexplained}$ in an RF ensemble instead of the minimum, or approximate an NN using a na\"ive surrogate grown with Gini impurity instead of our optimised method, the $\text{PercentUnexplained}$
is substantially worsened.

\begin{wrapfigure}{l}{2.75in}
    \centering
    \vspace{-0.4cm}
    \includegraphics[width=2.75in]{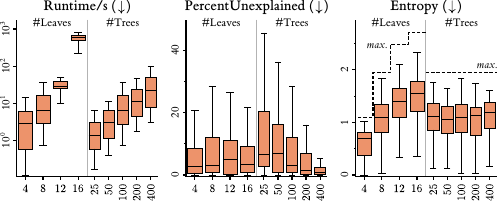}
    \vspace{-0.5cm}
    \caption{Impact of RF ensemble size on metrics.}
    \vspace{-0.2cm}
    \label{fig:model_size}
\end{wrapfigure}

\vspace{-0.275cm}
\paragraph{Impact of Model Size}

Figure \ref{fig:model_size} studies the effect of varying the number of trees and leaves per tree in an RF ensemble.
\ours{}'s runtime increases more dramatically with leaf count than tree count, matching expectations as theoretical complexity is exponential in the former but linear in the latter.
Fascinatingly, \textbf{the $\text{PercentUnexplained}$ tends to \textit{reduce} as the tree count increases}.
That is, in the trade-off between creating a more complex, harder-to-explain model and providing an increasing number of candidate explanations to select from, it appears the latter reliably dominates.
This is a very promising result for the scalability of \ours{} to large tree ensembles.
Explanation entropy increases with leaf count (i.e. number of conditional factors) but encouragingly tends to be a decreasing proportion of the theoretical maximum (dotted line).
Also note the metrics that \textit{don't} change: $\text{PercentUnexplained}$ is roughly constant across tree sizes (another positive scalablity result), and entropy is roughly constant across ensemble sizes.


\vspace{-0.275cm}
\paragraph{Comparison to Optimal Transport Baselines}

Prior shift explanation methods tackle somewhat different problems to ours, but with targeted additions, we can adapt two methods proposed by Kulinski \& Inouye \cite{kulinski2023explaining} as baselines.
Both begin by finding a pairing between data points from $P$ and $Q$ using optimal transport (OT).
The first method (OT-C) then groups the paired points into $k$ clusters and summarises the shift using a mean shift vector for each cluster.
Our addition is to then compute an attribution for each cluster with respect to the prediction shift of a target model $f$.
The second method
(OT-S) instead sparsifies the OT mapping, yielding a per-point shift vector that is nonzero along only $k$ features.
We then compute an SV for each nonzero feature to explain a prediction shift, which effectively combines the method with that proposed by Koebler et al. \cite{koebler2023towards}.
For both OT-C and OT-S, the hyperparameter $k$ dictates the number of factors in the explanation.
We set $k=7$ to match the number in our \ours{} explanations, enabling a like-for-like comparison.
Appendix~\ref{app:baseline_details} gives more details on our additions to both baselines.

\begin{figure*}[t]
    \centering
    \includegraphics[width=\textwidth]{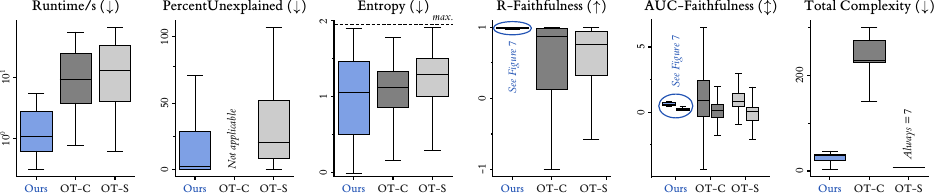}
    \vspace{-0.5cm}
    \caption{Quantitative comparison of \ours{} to adapted OT baselines on \textit{Folktables} data.
    }
    \vspace{-0.1cm}
    \label{fig:baseline_comparison_quant}
\end{figure*}

Figure~\ref{fig:baseline_comparison_quant} compares OT-C and OT-S to model-agnostic \ours{} for NN models on all $250$ \textit{Folktables} shifts.
\textbf{Both baselines have around $10\times$ higher runtime}.
While there is no meaningful notion of $\text{PercentUnexplained}$ for OT-C, we can compute it for OT-S, and find it to be clearly worse than our method, with a significance on the order of $p=10^{-19}$ under a one-sided Mann-Whitney U test.
The differences in entropy are less stark, but \ours{} still yields lower-entropy explanations with a significance of $p=0.04$ vs. OT-C and $p=10^{-7}$ vs. OT-S.
The baselines rarely achieve entropies much below $1$, indicating they rarely return attributions with just one or two large values.

To evaluate faithfulness for OT-C, we move data points by the mean shifts of their respective clusters, and for OT-S, we substitute in feature values from the corresponding paired points on a per-feature basis.
Querying model predictions for these perturbed points enables us to compute the R-Faithfulness and AUC-Faithfulness metrics.
Median R-Faithfulness is $0.875$ for OT-C and $0.755$ for OT-S, compared with $0.998$ for \ours{}.
The lower ends of the baseline distributions also drop below zero, indicating a strong misordering of attributions.
The magnitudes of AUC values tend to be higher for both baselines (indicating that perturbing points by shifting alters mean predictions more than reweighting), but more important is that the distributions of AUAC (left) and AUIAC (right) are proportionally less well-separated.
Taken together, these results indicate that \textbf{the OT-based explanations less faithfully reflect the effects of data interventions than \ours{}}.

As a final metric, we quantify the complexity of an explanation as the total number of numbers required to specify it in full.
OT-S explanations always have a complexity of just $7$ (one SV per nonzero feature) whereas OT-C also includes the means of all pre- and post-shift clusters, requiring up to $301$ numbers for the highest-dimensional \textit{Folktables} dataset.
\ours{}'s subgroup conditional explanations have an intermediate numerical complexity, typically between $20$ and $35$.

\begin{wrapfigure}{r}{2.5in}
    \centering
    \includegraphics[width=2.5in]{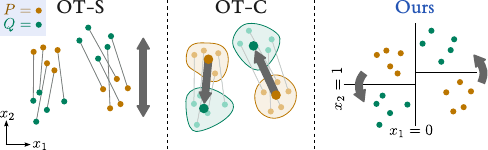}
    \vspace{-0.5cm}
    \caption{Qualitative shift representations used by our method and baselines.}
    \label{fig:baseline_comparison_qual}
\end{wrapfigure}

More qualitatively, we argue that \ours{} explanations achieve a stronger balance between interpretability and information content than the OT baselines.
As shown in Figure~\ref{fig:baseline_comparison_qual}, OT-S discards all information about a shift aside from the sparse subset of features (in this case $\{x_2\}$), which is simple but says nothing about \textit{how} those features have shifted.
OT-C represents a shift with vectors between clusters, each with the same dimensionality as the input space, which can be high for real-world problems.
The geometries of the clusters also have no simple descriptions.
Conversely, our subgroups are formed from the axis-aligned splits of decision trees, typically over just a few features, meaning their conditional probabilities can be described using an interpretable list of features and thresholds (e.g. the probability that $x_2<1$ given that $x_1<0$).





\vspace{-0.2cm}
\paragraph{Scaling to Large Trees}

\begin{wrapfigure}{r}{2.5in}
    \centering
    \vspace{-0.4cm}
    \includegraphics[width=2.5in]{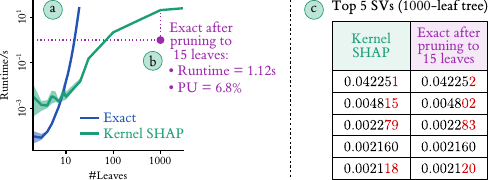}
    \vspace{-0.5cm}
    \caption{Comparison of Kernel SHAP and pruning methods for handling large trees.}
    \label{fig:scaling}
\end{wrapfigure}

A remaining concern one might have about \ours{} is how to achieve reasonable runtimes with large trees.
A simple way to address this, while accepting some SV approximation error, is to use the Kernel SHAP estimator \cite{lundberg2017unified}. Figure~\ref{fig:scaling}~\smash{\annot{a}} demonstrates this approach on \textit{Folktables} decision trees.
Kernel SHAP dramatically reduces the time complexity: explanations for $3000$-leaf trees are computed in the same time as the exact method for $19$-leaf trees.

Alternatively, the structure of our problem affords another way of handling large trees.
Since the largest SVs are almost always for conditionals of split nodes near the root of a tree (which impact the largest proportions of data), we can obtain very similar explanations by applying the exact method to a heavily-pruned version of the tree, which effectively serves
as a surrogate for the full one.
In practice, we create a pruned surrogate for a large tree by iteratively regrowing it from the root, at each step expanding the node with highest impurity as defined in Section~\ref{sec:model_agnostic}, and stopping when the desired size is reached.
Figure~\ref{fig:scaling}~\smash{\annot{b}} shows that applying \ours{} to a $15$-leaf pruned surrogate of a $1000$-leaf tree takes just $1.12$ seconds and achieves a $\text{PercentUnexplained}$ of $6.8\%$ with respect to the full tree.
Furthermore, \smash{\annot{c}} shows a very close agreement with the Kernel SHAP results. The top-five ranking of largest SVs is identical between the two methods, and there is numerical equality to at least four decimal places.
These results suggest that \textbf{each of these adaptations largely eradicates the main barrier to applying \ours{} to large models}.
See Appendix~\ref{app:large_trees} for more details.


\vspace{-0.1cm}
\section{Conclusion}

We propose \ours{}: an SV analysis of subgroup conditional probabilities, derived from tree structures, that provides simple, faithful and near-complete explanations of prediction shift across model classes.
Future work could explore using our model-agnostic method to explain
prediction shift in non-tabular domains given an interpretable semantic feature space \citep{de2020human}, such as one based on visual concepts  \cite{koh2020concept}.
We also
note that the model-agnostic method could be easily repurposed to explain mean shifts in any univariate function of data, widening its potential remit beyond the analysis of machine learning models.
We consider some limitations of \ours{} in Appendix~\ref{app:limitations}.


\newpage
\section*{Disclaimer}
This paper was prepared for informational purposes by the Artificial Intelligence Research group of JPMorgan Chase \& Co. and its affiliates ("JP Morgan'') and is not a product of the Research Department of JP Morgan. JP Morgan makes no representation and warranty whatsoever and disclaims all liability, for the completeness, accuracy or reliability of the information contained herein. This document is not intended as investment research or investment advice, or a recommendation, offer or solicitation for the purchase or sale of any security, financial instrument, financial product or service, or to be used in any way for evaluating the merits of participating in any transaction, and shall not constitute a solicitation under any jurisdiction or to any person, if such solicitation under such jurisdiction or to such person would be unlawful.

© 2025 JPMorgan Chase \& Co. All rights reserved.

\bibliography{bibliography}
\bibliographystyle{plain}


\newpage

\appendix

\section{Methodological Details and Discussion}

\subsection{Handling General Models via Scalarisation}
\label{app:scalarisation}

Although the notation in the main paper presents a model as a real-valued function $f:\mathcal{X}\rightarrow \mathbb{R}$, all we fundamentally require is a way of deterministically \textit{mapping} model predictions to real numbers.
This makes the framework considerably more general.
Formally, let $m:\mathcal{X}\rightarrow\mathcal{Y}$ be a general model that maps inputs to an output space $\mathcal{Y}$.
We need assume no particular structure on $\mathcal{Y}$, provided we can also define a \textit{scalarisation} function $s:\mathcal{Y}\rightarrow \mathbb{R}$.
The composite $f=m\circ s$ is a real function as required.
A selection of examples are given below:
\begin{itemize}
    \item If $f$ is a univariate regression model, we can just use the identity scalarisation $s(y) \coloneqq y$.
    \item If $f$ is a $k$-class classifier (i.e. $\mathcal{Y}=\{0,\dots,k-1\}$), we can define $s(y) \coloneqq \mathbb{I}[f(x)\in Y^*]$ for some set of target classes $Y^*\subset \mathcal{Y}$ (which may be a singleton).
    Since the range of the indicator function ($\{0,1\}$) is a subset of $\mathbb{R}$, our method can handle it without modification.
    In this case, $\mu_P$ and $\mu_Q$ can be understood as the probabilities of $f$ predicting a target class under distributions $P$ and $Q$ respectively.
    \begin{itemize}
        \item Several experiments in the paper consider binary classification models.
        For these, we simply define $Y^* = \{1\}$.
    \end{itemize}
    \item The preceding case can be generalised to any other kind of output space (e.g. $\mathcal{Y}=\mathbb{R}^d$ for multivariate regression) given a subset of interest $Y^*\subset \mathcal{Y}$.
    \item If the model predicts distributions over the output space $f:\mathcal{X}\rightarrow\Delta(\mathcal{Y})$ (probabilistic prediction), or sets $f:\mathcal{X}\rightarrow 2^\mathcal{Y}$ (conformal prediction) rather than single points, we can define $s$ as a scalar measure of predictive uncertainty, such as entropy for probabilistic prediction or the size of the predictive set for conformal prediction. In this case, $\mu_P$ is the model's average uncertainty under the distribution $P$.
\end{itemize}

\subsection{Justification for Use of Subgroup Conditional Probabilities Instead of Joint Probabilities}
\label{app:why_subgroup}

We identify a mixture of theoretical, practical and empirical reasons for favouring an analysis of subgroup conditionals.

\paragraph{Reason \# 1: Avoiding Renormalisation}

Consider an input space consisting of two binary features $x_1$ and $x_2$.
Considering the two features jointly, there are four possible subgroups: $(x_1, x_2)$, $(x_1, \neg x_2)$, $(\neg x_1, x_2)$ and $(\neg x_1, \neg x_2)$, where ``$x_i$'' is used as shorthand for ``$x_i=\ $\textit{True}'' and ``$\neg x_i$'' means ``$x_i=\ $\textit{False}''.
Suppose that we have two distributions $P$ and $Q$ such that
\begin{itemize}
    \item $P(x_1, x_2)=0.1$, $P(x_1, \neg x_2)=0.3$, $P(\neg x_1, x_2)=0.4$, $P(\neg x_1, \neg x_2)=0.2$;
    \item $Q(x_1, x_2)=0.3$, $Q(x_1, \neg x_2)=0.2$, $Q(\neg x_1, x_2)=0.2$, $Q(\neg x_1, \neg x_2)=0.3$.
\end{itemize}
To compute SVs for these joint probabilities, we would need to measure the effect (on the mean prediction of a model target $f$) of swapping their values from $P$ to $Q$.
Imagine we are trying to swap from $P(x_1, x_2)=0.1$ to $Q(x_1, x_2)=0.3$ as part of such an analysis.
We encounter a problem: performing this swap in isolation creates an invalid distribution.
The probabilities of all subgroups must sum to $1$, but  $Q(x_1, x_2) + P(x_1, \neg x_2) + P(\neg x_1, x_2) + (\neg x_1, \neg x_2) = 1.2$.
We therefore face an unavoidable requirement to \textit{renormalise} the other three probabilities to make the distribution valid.
There are infinitely many ways in which this could be done, each of which would produce a different set of final SV attributions, and we are aware of no good reasons for preferring one over the other.
This makes the problem of computing SVs for joint probabilities fundamentally ill-posed.

In contrast, suppose that we factorise the distributions into conditional probabilities:
\begin{itemize}
    \item $P(x_1)=0.4$, $P(x_2|x_1)=0.25$, $P(x_2|\neg x_1)=0.333$;
    \item $Q(x_1)=0.5$, $Q(x_2|x_1)=0.6$, $Q(x_2|\neg x_1)=0.4$.
\end{itemize}
Computing SVs for these conditional probabilities would again swapping their values from $P$ to $Q$.
Imagine we are trying to swap from $P(x_2|x_1)=0.25$ to $Q(x_2|x_1)=0.6$ as part of such an analysis.
This can now be done unproblematically, with no need to change either of the other two conditional probabilities to maintain a valid distribution.
This means that, given a chosen factorisation of $P$ and $Q$ into conditional probabilities, the problem of computing SVs for conditional probabilities is well-posed.
In the paper, we derive our conditional factorisations from decision trees.

\paragraph{Reason \# 2: Usefulness of Explanations}

More informally, we hypothesise that subgroup conditional SVs provide a more interpretable and useful form of explanation than ones based on joint probabilities.
To illustrate this, let us set aside the aforementioned renormalisation issue and suppose that we have computed an SV for the change in joint probability $P(x_1, x_2)\rightarrow Q(x_1, x_2)$.
Furthermore, suppose that this SV is a large positive value, implying that the change has had a large positive effect on the mean prediction of a target model $f$.
Attempting to interpret this result, we are left unclear as to whether the change has involved an increase in the probability of $x_1$ while $x_2$ has remained unchanged, an increase in $x_1$ with $x_2$ unchanged, or a mixture of increases in both probabilities.
In contrast, if instead we are presented with a large SV for the specific conditional $(x_2|x_1)$, we know precisely the structure of the change that has occurred: among inputs where $x_1=\ $\textit{True}, there has been an increase in the proportion for which $x_2$ is also \textit{True}.
We believe that this more specific information is likely to be more useful for most downstream applications, including being more actionable for interventions that model developers may want to make on the data pipeline or the model itself. 

\paragraph{Reason \# 3: Counterintuitive Results}

\begin{figure}[H]
    \centering
    \includegraphics[width=\textwidth]{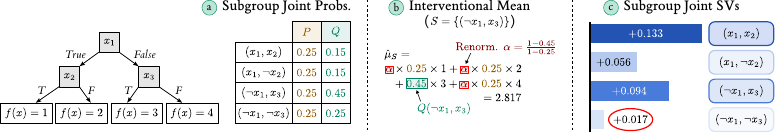}
    \caption{Illustration of a counterintuitive result when performing the SV analysis using subgroup joint probabilities.}
    \label{fig:single_tree_illustrative_with_joint}
\end{figure}

The joint SV analysis can be instantiated for decision trees by swapping the probabilities of each leaf node from $P$ to $Q$, rather than the conditional probabilites at each split node, and choosing a renormalisation scheme (note that this choice is arbitrary; see above).
For demonstrative purposes, we renormalise the probabilities of all leaves not included in the intervention by scaling them uniformly by a factor $\alpha$ such that the sum of all leaf probabilities remains at $1$.
In Figure \ref{fig:single_tree_illustrative_with_joint}, we apply this method to the toy example used in the main paper (albeit with a slightly different $Q$ distribution).
\smash{\annot{a}} shows the subgroup joint probabilities associated with each of the four leaves, \smash{\annot{b}} illustrates the renormalisation process and resultant mean calculation for intervention $S=\{(\neg x_1, x_3)\}$, and \smash{\annot{c}} shows the final results of the subgroup joint SV analysis.
The SVs for all four joint factors are positive, which implies that they have a positive impact on the prediction shift from $P\rightarrow Q$.
Crucially, however, this includes $(\neg x_1, \neg x_3)$, which receives an SV of $+0.017$ despite the fact that its probability remains constant at $0.25$ for both $P$ and $Q$.
This violates the basic intuition that if a factor does not change between two distributions, it should receive an attribution of zero with respect to any prediction shift.
With an SV analysis based on subgroup conditionals, such violations never occur.

\paragraph{Reason \# 4: Empirical Performance}

A final, less fundamental, reason for preferring subgroup conditional (split-level) factors to joint (leaf-level) factors is that they seem to give better empirical results under some of our evaluation metrics.
In preliminary experiments on \textit{Folktables}, we found that conditional SVs achieved somewhat lower $\text{PercentUnexplained}$ values on average.
The joint method also requires marginally higher runtime: since a tree always has one more leaf node ($4$ in the example above) than its number of split nodes ($3$), this creates one more factor in the SV analysis.

\newpage

\subsection{Pseudocode for Subgroup Conditional Shapley Value Calculation (Single Decision Tree)}
\label{app:pseudocode}

\renewcommand{\algorithmicrequire}{\textbf{Inputs:}}
\renewcommand{\algorithmicprocedure}{\textbf{def}}
\begin{algorithm}[H]
    \caption{Calculation of subgroup conditional Shapley values.}
    \begin{algorithmic}[1]
        \Require Distributions $P$ and $Q$, subgroup conditionals $C$, tree leaves $L$, per-leaf predictions $F$;
        \State $\hat{\mu} \gets \{\}$
        \For{$S\subseteq C$}
            \State $\hat{\mu}[S] \gets 0$
            \For{$l \in L$}
                \State $z = \textsc{GetInterventionalLeafProb}(P, Q, C, S, l)$  \Comment{See Algorithm \ref{alg:intleafprob} below.}
                \State $\hat{\mu}[S] \gets \hat{\mu}[S] + (z \times F[l])$
            \EndFor
        \EndFor
        \State $\phi \gets \{\}$
        \For{$c \in C$}
            \State $\phi[c] \gets 0$
            \For{$S\subseteq C \setminus \{c\}$}
            \State $w \gets \frac{|S|!(|C| - |S| - 1)!}{|C|!}$
                \State $\phi[c] \gets \phi[c] + (w \times (\hat{\mu}[S\cup\{c\}] - \hat{\mu}[S]))$
            \EndFor
        \EndFor
        \State \Return $\phi$
    \end{algorithmic}
\end{algorithm}
\begin{algorithm}[H]
    \caption{Calculation of the probability of leaf $l \in L$ under subgroup conditional interventions $S\subseteq C$.}
    \label{alg:intleafprob}
    \begin{algorithmic}[1]
        \Procedure{GetInterventionalLeafProb}{$P, Q, C, S, l$}
            \State $z \gets 1$
            \For{$(g_1|g_2) \in C$} \Comment{Each $c\in C$ is a subgroup conditional $(g_1|g_2)$.}
                \If{$l \subseteq g_2$} \Comment{Check if this leaf is affected by this subgroup conditional.}
                    \If{$(g_1|g_2) \in S$} \Comment{If this subgroup conditional is in the intervention set, use probabilities from $Q$.}
                        \If{$l \subseteq g_1$}
                            \State $z \gets z \times Q(g_1|g_2)$
                        \Else
                            \State $z \gets z \times (1 - Q(g_1|g_2))$
                        \EndIf
                    \Else \Comment{Otherwise, use probabilities from $P$.}
                        \If{$l \subseteq g_1$}
                            \State $z \gets z \times P(g_1|g_2)$
                        \Else
                            \State $z \gets z \times (1 - P(g_1|g_2))$
                        \EndIf
                    \EndIf
                \EndIf
            \EndFor
            \State \Return $z$
        \EndProcedure
    \end{algorithmic}
\end{algorithm}

\subsection{Causal Graph Interpretation} \label{app:causal_graph}

\begin{figure}[H]
    \centering
    \includegraphics[width=\textwidth]{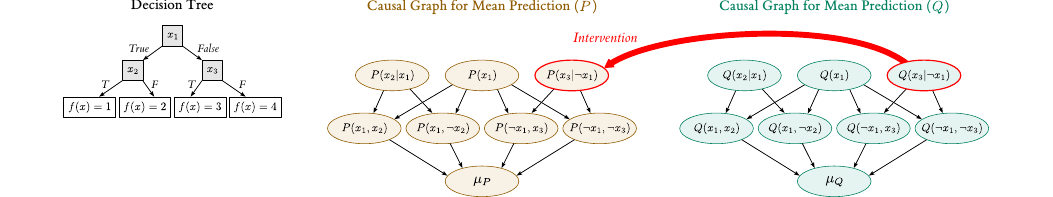}
    \caption{Interpretation of our method as interventions on a causal graph.}
    \label{fig:tree_to_causal_model}
\end{figure}

Several prior works construct SV explanations of distribution shifts by swapping conditional probabilities in a known causal graph from their values under $P$ to their values under $Q$ \citep{budhathoki2021why, zhang2022why}.
Our method bears some resemblance to this approach, as the mean prediction of a decision tree can be given a kind of causal graph representation, where the nodes are conditional and joint subgroup probabilities (see Figure~\ref{fig:tree_to_causal_model}).
Swapping a conditional probability from $P$ to $Q$, as we do in our SV analysis, is effectively a causal intervention on one or more of the conditional nodes.

Since the conditional nodes have no ancestors in the causal graph, the approach may also be theoretically related to the distal variant of the asymmetric SVs framework \citep{frye2020asymmetric}.
The alternative approach based on intervening on joint (leaf) probabilities, discussed in Appendix \ref{app:why_subgroup}, resembles the proximate variant.

\subsection{Computing a Separate Shapley Value for Each Leafwise Mean}
\label{app:per_leafwise_mean}

In the main paper, we describe how a single extra SV $\phi_{\text{LeafMeans}}$ can be used to quantify the remaining ensemble prediction shift left unexplained for by the subgroup conditional probabilities of any single tree, which is due to a change in all leafwise mean predictions.
An alternative valid approach would be to compute a separate SV for \textit{each} leafwise mean.

Letting $L$ denote the set of all leaves in the tree, this would involve computing an interventional mean $\hat{\mu}_S$ for each subset of interventions $S\subseteq C^+$, where $C^+ = C \cup \{\text{LeafMean}[l]:l \in L\})$ and the $\text{LeafMean}[l]$ intervention swaps \textit{only} the leafwise mean of leaf $l \in L$ from its value under $P$ (i.e. $\mathbb{E}_{x\sim P}[f(x)|x\in l]$) to its value under $Q$ (i.e. $\mathbb{E}_{x\sim Q}[f(x)|x\in l]$).
In turn, the SV for each leafwise mean would be defined as
\[
    \phi_{\text{LeafMean}[l]} \coloneqq \sum_{S\subseteq (C^+ \setminus \text{LeafMean}[l])}\frac{|S|!(|C^+|-|S|-1)!}{|C^+|!}\Big[\hat{\mu}_{S\cup\{\text{LeafMean}[l]\}} - \hat{\mu}_{S}\Big].
\]

One appeal of this alternative is that it enables more granular insight into \textit{where} (i.e. which leaves) the unexplained prediction shift is coming from: the leaves with large SVs.
This might enable improved evaluation metrics or even be of direct interest to an end user.
However, by expanding the additional interventions from just one to $|L|$, it causes an exponential increase in computational cost.
We are yet to pursue it further for this reason, although we mention it in case it inspires future work.



\subsection{Occasional Explanation Failure Mode}
\label{app:failure_mode}

In Figure~\ref{fig:ensemble_experiment}
of the main paper, we annotate a small number of trees in an ensemble for which no explanation can be computed.
This rare failure mode occurs when exactly zero data points in the empirical dataset for either $P$ and $Q$ reach a split node, leading the associated conditional probability to be undefined. In turn, we cannot perform an intervention on this conditional, and thus cannot complete the SV analysis.
In our experiments, we never encountered an ensemble model in which more than a handful of trees encountered this issue, so it was always possible to return an explanation by selecting the tree with lowest $\text{PercentUnexplained}$.
We also emphasise that this issue only arises when explaining tree-based models using their own conditional structure. In the model-agnostic setting, we have direct control of how the surrogate tree is grown, so add an explicit constraint that prevents undefined conditionals from ever occurring.

\subsection{Justification of Proposed Impurity Function as a Proxy for $\text{PercentUnexplained}$}
\label{app:impurity_as_proxy}

When growing a surrogate tree, our ultimate objective is to minimise the $\text{PercentUnexplained}$ by minimising the absolute value of the residual SV $\phi_\text{LeafMeans}$.
This value is defined using the standard SV formula:
\[
|\phi_{\text{LeafMeans}}| \coloneqq \Big|\sum_{S\subseteq C} w(S) \Big[\hat{\mu}_{S\cup\{\text{LeafMeans}\}} - \hat{\mu}_{S}\Big]\Big|,
\]
where $w(S)=\frac{|S|!(|C\cup \{\text{LeafMeans}\}|-|S|-1)!}{|C\cup \{\text{LeafMeans}\}|!}$ and $\hat{\mu}_S$ is the mean prediction of the target model $f$ under the distribution defined by the set of interventions $S\subseteq (C\cup\{\text{LeafMeans}\})$.
We can expand this expression as follows:
\[
|\phi_{\text{LeafMeans}}| \coloneqq  \Big|\sum_{S\subseteq C} w(S) \sum_{l\in L} Z_S(l) \Big[\mathbb{E}_{x\sim Q}[f(x)|x\in l] - \mathbb{E}_{x\sim P}[f(x)|x\in l]\Big]\Big|
\]
\[
= \Big|\sum_{l\in L} \Big(\sum_{S\subseteq C} w(S) Z_S(l)\Big) \times \Big[\mathbb{E}_{x\sim Q}[f(x)|x\in l] - \mathbb{E}_{x\sim P}[f(x)|x\in l]\Big]\Big|,
\]
where $L$ is the set of leaves in the surrogate and $Z_S(l)\in[0, 1]$ is the probability of leaf $l\in L$ under interventions $S\subseteq C$.
This probability is computed using the \textsc{GetInterventionalLeafProb} function defined in Appendix~\ref{app:pseudocode}.

Recall that a surrogate tree is grown by splitting an extant leaf $l$ into two child leaves $l_\text{leaf}\cup l_\text{right}=l$ to minimise their total impurity $I(l_\text{left})+I(l_\text{right})$, where impurity is a local function of each child leaf.
The absolute value expression above cannot be written as a sum of local leaf impurities, but it is upper-bounded by an expression with the absolute value operator inside the summation, that \textit{can} be written as a sum of impurities:
\[
|\phi_{\text{LeafMeans}}| \coloneqq \Big|\sum_{l\in L} \Big(\sum_{S\subseteq C} w(S) Z_S(l)\Big) \times \Big[\mathbb{E}_{x\sim Q}[f(x)|x\in l] - \mathbb{E}_{x\sim P}[f(x)|x\in l]\Big]\Big|\ \ \ \ \ \leq\ \ \ \ \  \sum_{l\in L} I^*(l),
\]
\[
\text{where}\ \ \ \ \ I^*(l) \coloneqq \Big(\sum_{S\subseteq C} w(S) Z_S(l)\Big) \times \Big|\mathbb{E}_{x\sim P}[f(x)|x\in l] - \mathbb{E}_{x\sim Q}[f(x)|x\in l]\Big|.
\]

Hence, an optimal surrogate tree growth algorithm would use $I^*(l_\text{left})+I^*(l_\text{right})$ as the impurity to minimise during splitting.
However, evaluating the bracketed quantity $\sum_{S\subseteq C} w(S) Z_S(l)$ for every candidate $(l_\text{left}, l_\text{right})$ pair would explode in complexity as tree growth proceeds.
It is exponential in the number of conditionals (hence leaves), and effectively demands the full SV calculation to be completed in the loop of tree growth.
We need to replace this quantity with a more tractable proxy.

To do so, we note that $\sum_{S\subseteq C} w(S) Z_S(l)$ is a weighted average probability of leaf $l$ under all sets of interventions.
We hypothesise that in general, this probability will fall somewhere close to midway between the probability under distribution $P$ (which is the distribution when $S=\emptyset$) and the probability under distribution $Q$ (when $S=C$).
That is, we hypothesise that
\[
    \sum_{S\subseteq C} w(S) Z_S(l) \approx \frac{P(l) + Q(l)}{2}.
\]
We make no universal claim about the accuracy of this approximation, as one can adversarially construct a tree structure and distributions $P$ and $Q$ for which it is arbitrarily inaccurate.
However, we can use numerical simulation to show it holds in expectation under simple distribution assumptions.

We consider a decision tree of depth $3$, with $8$ leaves and $7$ split nodes.
For each split node, we uniform-randomly sample two probabilities $\in[0,1]$ of the test at that node evaluating to \textit{True}, one for $P$ and one for $Q$.
We use these probabilities to compute $P(l)$ and $Q(l)$ for each leaf.
We then iterate over all possible interventions to swap subsets of split node probabilities from $P$ to $Q$, and thereby compute $\sum_{S\subseteq C} w(S) Z_S(l)$ for each leaf.
We repeat this entire process $5000$ times.

\begin{figure}[H]
    \centering
    \includegraphics[width=\textwidth]{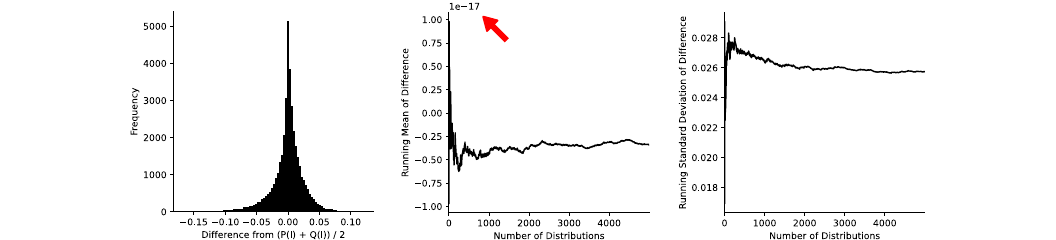}
    \caption{Empirical demonstration that $\sum_{S\subseteq C} w(S) Z_S(l) \approx \frac{P(l) + Q(l)}{2}$ in expectation.}
    \label{fig:approx_error}
\end{figure}

The left subplot of Figure~\ref{fig:approx_error} shows the distribution of differences between $\sum_{S\subseteq C} w(S) Z_S(l)$ and $\frac{P(l) + Q(l)}{2}$ across all leaves.
It is tightly clustered around zero, seemingly with a double exponential form.
The middle and right subplots show how the mean and standard deviation of this distribution converge as we iterate over the $5000$ random $(P, Q)$ pairs.
The mean converges to almost exactly zero (note y-axis scale) and the standard deviation converges to $0.026$.
The weighted average interventional probability lies within a margin of $0.05$ from $\frac{P(l) + Q(l)}{2}$ $93.4\%$ of the time, and within a $0.1$ margin $99.4\%$ of the time.

We also try sampling probabilities for $P$ and $Q$ in a correlated way, and find that the distribution becomes even more concentrated around zero as the correlation increases.
We can therefore be confident that our hypothesis should hold for most distributions, provided they are not adversarially constructed.

This confidence in our hypothesis permits us to make the following approximate substitution:
\[
|\phi_{\text{LeafMeans}}| \lessapprox \sum_{l\in L} I(l),
\]
\[
\text{where}\ \ \ \ \ I(l) \coloneqq \Big(\frac{P(l) + Q(l)}{2}\Big) \times \Big|\mathbb{E}_{x\sim P}[f(x)|x\in l] - \mathbb{E}_{x\sim Q}[f(x)|x\in l]\Big|,
\]
which is a much more tractable impurity function.
Therefore, minimising $I(l_\text{left})+I(l_\text{right})$ when splitting should provide a good proxy for minimising $|\phi_{\text{LeafMeans}}|$ and, by extension, the $\text{PercentUnexplained}$.
Note that the impurity expression given in the main paper drops the factor of $\frac{1}{2}$ for brevity.

\section{Dataset and Experiment Details}
\label{app:dataset_and_experiment_details}

All experiments in this paper are run on a single \texttt{t3.2xlarge} AWS instance, which is a 2.5GHz 8 vCPU Linux machine with 32GB of memory.
Our code is implemented in Python 3.8.13 and relies on only standard libraries for data science and machine learning, specifically \texttt{numba==0.58.1}, \texttt{numpy==1.24.4}, \texttt{pandas==2.0.3}, \texttt{scikit\_learn==1.3.2}, \texttt{scipy==1.10.1} and \texttt{xgboost==2.0.3}.
We use explicit seeding in all experiments to ensure reproducibility.
Our code is included in the supplemental material.

\subsection{Illustrative Experiments on Benchmark Datasets}
\label{app:expt_illustrative}

For the experiments accompanying the methods sections, we simulate shifts in popular benchmark datasets by partitioning them into two halves with distinct distributions.
Specifically, for each dataset, we choose a single partitioning feature and threshold, and then define distributions $P$ and $Q$ as all data with feature values below and above that threshold respectively (or vice versa).
We then drop the partitioning feature from the dataset and train a model (either a decision tree, random forest or neural network) on all data from both $P$ and $Q$.
Finally, we take the change in the trained model's mean prediction on $P$ ($\mu_P$) and the mean prediction on $Q$ ($\mu_Q$) as the prediction shift to be explained.
Specific details on each dataset are given below.

\paragraph{California House Prices}

This dataset was first featured in \cite{pace1997sparse} and is available (at the time of writing) at \texttt{https://www.kaggle.com/datasets/camnugent/california-housing-prices}.
It concerns the prediction of the average 1990 price of Californian houses in a residential `block group'
(a regression task) given eight numerical features.
As the partitioning feature, we choose the average income of houses in the block, and split the dataset at the median ($\$35,348$), so that $P$ contains houses in low-income areas and $Q$ contains houses in high-income areas.
The model for this task is a \texttt{scikit-learn} \texttt{DecisionTreeRegressor} with \texttt{max\_leaf\_nodes=8} and otherwise default hyperparameters.

\paragraph{Breast Cancer Wisconsin (Original)}

This dataset was first featured in \cite{mangasarian1990cancer} and is available (at the time of writing) at \texttt{https://archive.ics.uci.edu/dataset/15/} \texttt{breast+cancer+wisconsin+original}.
It concerns the prediction of whether a breast mass is benign or malignant (a classification task) given nine numerical cytological characteristics computed from images.
As the partitioning feature, we choose the date of reporting, and define $P$ as cases reported in January 1989 (just over half the total) and $Q$ as cases reported at later dates up to November 1991.
The target model for this task is a \texttt{scikit-learn} \texttt{RandomForestClassifier} with \texttt{n\_estimators=100}, \texttt{max\_leaf\_nodes=8}, \texttt{random\_state=0} and otherwise default hyperparameters.

\paragraph{Home Equity Line of Credit}

This dataset was first featured in \cite{heloc} and is available (at the time of writing) at \texttt{https://www.kaggle.com/datasets/averkiyoliabev/} \texttt{home-equity-line-of-credit}.
It concerns the prediction of whether a client repaid a home equity loan within two years (a classification task) given $23$ numerical features.
As the partitioning feature, we choose \texttt{ExternalRiskEstimate}, which uses several risk markers to give a score (higher means lower risk).
We partition at the median value of $72.0$, defining $P$ as clients with scores above the median (low risk) and $Q$ and clients with scores below (high risk).
The target model for this task is a \texttt{scikit-learn} \texttt{MLPClassifier} with \texttt{random\_state=0} and otherwise default hyperparameters.

\paragraph{Pima Indians Diabetes}

This dataset was first featured in \cite{smith1988using} and is available (at the time of writing) at \texttt{https://www.kaggle.com/datasets/uciml/pima-indians-diabetes-database}.
It concerns the prediction of whether or not a patient has diabetes (a classification task) given eight numerical diagnostic measurements.
As the partitioning feature, we choose \texttt{DiabetesPedigreeFunction}, which estimates a patient's diabetes risk based on family history (higher means higher risk).
We partition at the median value of $0.3725$, defining $P$ as patients with estimates below the median (low risk) and $Q$ and patients with estimates above (high risk).
The target model for this task is a \texttt{scikit-learn} \texttt{MLPClassifier} with \texttt{random\_state=0} and otherwise default hyperparameters.

\subsection{Quantitative Evaluation on \textit{Folktables}}
\label{app:expt_folktables}

For our large-scale quantitative  experiments, we use five datasets from the \textit{Folktables} repository of US Census datasets \cite{ding2021retiring}, which is available (at the time of writing) at \texttt{https://github.com/socialfoundations/folktables}.
The five datasets are \textit{ACSIncome} (predict whether an individual's income is above $\$50k$), \textit{ACSPublicCoverage} (predict whether an individual is covered by public health insurance), \textit{ACSMobility} (predict whether an individual had the same residential address one year ago), \textit{ACSEmployment} (predict whether an individual is employed) and \textit{ACSTravelTime} (predict whether an individual has a commute to work longer than 20 minutes).
All of these are binary classification tasks.

Each dataset is partitioned into subsets for all $50$ US states (plus Puerto Rico) and five years from $2014$ to $2018$ inclusive, effectively creating $51\times 5=255$ unique distributions.
We can simulate a distribution shift by choosing one of these state-year distributions (e.g. Texas in 2015) as $P$ and another (e.g. Michigan in 2017) as $Q$.
This enables a potential $(255\times 254)/2=32,385$ unique distribution shifts per dataset (ignoring mirror-images), of which we sample $50$ at random for our experiments.
We use these same $50$ shifts for all five datasets, creating a total of $250$ shifts over which evaluation metrics are aggregated.

For each of these $250$ shifts, we train a model on all data from both $P$ and $Q$.
We then take the change in the trained model's mean prediction on $P$ ($\mu_P$) and the mean prediction on $Q$ ($\mu_Q$) as the prediction shift to be explained.
Our main results in Figure~\ref{fig:headline_results} of the paper use the following models (with default hyperparameters unless otherwise specified):
\begin{itemize}
    \item \texttt{scikit-learn} \texttt{DecisionTreeClassifier} (\texttt{max\_leaf\_nodes=8}).
    \item \texttt{scikit-learn} \texttt{RandomForestClassifier} (\texttt{n\_estimators=100,\\max\_leaf\_nodes=8,random\_state=0}).
    \item \texttt{xgboost} \texttt{XGBClassifier} (\texttt{n\_estimators=100,max\_leaves=8,random\_state=0}).
    \item \texttt{scikit-learn} \texttt{MLPClassifier} with \texttt{random\_state=0}.
    Under the default hyperparameters, this model has one hidden layer of $100$ neurons and ReLU activations.
    It uses the Adam optimiser with a learning rate of $0.001$.
    We learn a surrogate tree with $8$ leaves for this model.
\end{itemize}

The model size experiment in Figure~\ref{fig:model_size} of the paper studies the effect of the \texttt{max\_leaf\_nodes} and \texttt{n\_estimator} hyperparameters of \texttt{RandomForestClassifier} on our method's performance.
We vary \texttt{max\_leaf\_nodes} in $\{4, 8, 12, 16\}$ and \texttt{n\_estimators} in $\{25, 50, 100, 200, 400\}$, keeping the other hyperparameter fixed in each case.


\subsection{Demonstration of Scaling to Large Trees}
\label{app:large_trees}

We explore two strategies for scaling the method to large trees:

\begin{itemize}
    \item \textbf{Kernel SHAP:} We achieve this through a trivial integration of our method with the \texttt{KernelExplainer} class from the \texttt{shap==0.46.0} Python package \cite{lundberg2017unified}, which we instantiate with default hyperparameters.
    \item \textbf{Pruned Surrogate:} To generate a pruned surrogate of a large tree, we begin by pruning it back until it has just two leaves, denoted by $l_1$ and $l_2$. We compute the impurity of these two leaves, $I(l_1)$ and $I(l_2)$, where $I$ is the impurity function defined in Section~\ref{sec:model_agnostic} of the main paper.
    We then \textit{expand} whichever of $l_1$ and $l_2$ has the higher impurity, which means reinstating its two child nodes from the full tree. This results in a tree with three leaves. We repeat this process, expanding one node at a time based on the impurity criterion, until a desired tree sized is reached. We then apply our Shapley value method to this pruned tree as a surrogate for the full one, including computing the extra SV $\phi_{\text{LeafMeans}}$ to account for shift in the full tree's mean prediction within each leaf of the pruned one.
\end{itemize}

We apply these techniques to large decision trees for the \textit{ACSPublicCoverage} dataset from \textit{Folktables}.
We define distributions $P$ and $Q$ as data from different arbitrary mixtures of states.
The runtime results in Figure~\ref{fig:scaling} are averaged over five random seeds.
For the $1000$-leaf tree that we explain using both Kernel SHAP and a pruned surrogate, $\mu_P=0.29$ and $\mu_Q=0.35$, for an overall prediction shift of $\mu_Q-\mu_P=+0.06$. The largest SV of $+0.042$ is associated with the conditional probability that $(\text{state}=\text{Alaska}\ |\ \text{disabled})$, which decreases from $78\%$ in $P$ to $29\%$ in $Q$.
That is, the tree predicts a higher rate of public health coverage in distribution $Q$ because fewer of the disabled people live in Alaska.
This indicates that the model has identified and encoded a geographic inequality in public health provision for individuals with disabilities.

\section{Faithfulness Metrics}
\label{app:faithfulness_metrics}

To evaluate the faithfulness of our method's explanations, we study the effect of reweighting data from $P$ to match the subgroup conditional probabilities from $Q$ and vice versa.
For example, suppose that our method has returned an SV $\phi_{(g_1|g_2)}$ for some subgroup conditional $(g_1|g_2)$ associated with a split node in a (surrogate) tree.
We define the following reweighting function:
\[
    \lambda_{(g_1|g_2)}(x) \coloneqq \left\{\begin{array}{ll}
       \left\{\begin{array}{ll}
            \frac{Q(g_1|g_2)}{P(g_1|g_2)} & \text{if }x \in g_1,\vspace{0.15cm} \\
            \frac{1 - Q(g_1|g_2)}{1 - P(g_1|g_2)} & \text{otherwise},
       \end{array}\right.
       & \text{if }x \in g_2,\vspace{0.15cm} \\
       1                 & \text{otherwise}.
    \end{array}\right.
\]
In turn, we can compute the mean prediction of the target model $f$ under a reweighted distribution of samples from $P$:
\[
    \tilde{\mu}_{(g_1|g_2)} \coloneqq \mathbb{E}_{x\sim P}[f(x) \times  \lambda_{(g_1|g_2)}(x)].
\]
This basic idea of computing mean predictions under a reweighted distribution forms the basis of two faithfulness metrics.

\paragraph{R-Faithfulness}

The intuition behind this metric, inspired by prior work \cite{bhatt2021evaluating,koebler2023towards}, is that the reweighted mean $\tilde{\mu}_{(g_1|g_2)}$ should be higher if the SV $\phi_{(g_1|g_2)}$ is higher.
To formalise this, let $C$ be the set of all subgroup conditionals for which SVs have been computed, of which $(g_1|g_2)$ is one element.
We define R-Faithfulness as the Pearson correlation between the SVs and their corresponding reweighted means:
\[
    \text{R-Faithfulness} \coloneqq corr\Big((\phi_c,\ \forall\ c\in C), (\tilde{\mu}_c,\ \forall\ c\in C)\Big).
\]
We also complete the entire process in a reversed way, reweighting data from $Q$ to match the conditional probabilities from $P$, computing reweighted means, and measuring the correlation with the SVs (which in this case needs to be flipped).
All box plots in Figure~\ref{fig:headline_results} of the paper aggregate R-Faithfulness values over both `forward' reweightings $P\rightarrow Q$ and `backward' reweightings $Q\rightarrow P$.
We find the forward and inverse directions give similar results in practice.

\paragraph{AUC-Faithfulness}

As a generalisation of reweighting data to modify one conditional probability at a time, we can compute a cumulative weight to modify a set of conditionals $S \subseteq C$ and measure the reweighted mean prediction:
\[
    \tilde{\mu}_{S} \coloneqq \mathbb{E}_{x\sim P} \Big[ f(x) \times \prod_{c\in S} \lambda_{c}(x) \Big].
\]

The intuition behind the AUC-Faithfulness metric is that if we add conditionals to this cumulative reweighting analysis in descending order of their SVs, it should have a decreasing effect on the mean prediction.
Conversely, if we cumulatively reweight in ascending SV order, it should have an increasing effect.
This would indicate that the SVs are faithfully ordered.

Formally, let $C^{\downarrow}$ be a list that sorts the conditionals $C$ in descending SV order in the direction of the prediction shift, i.e. $\phi_{C^{\downarrow}[i]} \leq \phi_{C^{\downarrow}[i - 1]}$ if $\mu_Q - \mu_P \geq 0$, and $\phi_{C^{\downarrow}[i]} \geq \phi_{C^{\downarrow}[i - 1]}$ otherwise.
Furthermore, let $C^{\uparrow}$ denote the opposite order.
Adopting language from prior work \citep{schnake2021higher,kariyappa2024progressive}, we define the area under the activation curve (AUAC) and area under the inverse activation curve (AUIAC) as follows:
\[
    \text{AUAC} \coloneqq \frac{1}{|C^{\downarrow}|}\sum_{i=1}^{|C^{\downarrow}|}\frac{\tilde{\mu}_{C^\downarrow[1:i]} - \mu_P}{\mu_Q - \mu_P};\ \ \ \ \ \text{AUIAC} \coloneqq \frac{1}{|C^{\uparrow}|}\sum_{i=1}^{|C^{\uparrow}|}\frac{\tilde{\mu}_{C^\uparrow[1:i]} - \mu_P}{\mu_Q - \mu_P}.
\]
If the AUAC exceeds the AUIAC, then we can say that reweighting in descending SV order has a decreasing effect on the mean prediction and vice versa.
By plotting the distribution of AUAC and AUIAC values across many shifts and measuring how consistently the AUAC exceeds the AUIAC using a statistical significance test, we obtain an overall faithfulness metric.

As with R-Faithfulness, we compute AUAC and AUIAC values for backward reweightings $Q\rightarrow P$ alongside the forward reweightings $P\rightarrow Q$ assumed in the equations above.
The box plots in Figure~\ref{fig:headline_results} again aggregate over both directions.

\section{Baseline Details}
\label{app:baseline_details}

We base our implementations of OT-C and OT-S on the official code repository released by Kulinski\cite{kulinski2023explaining}, which is available (at the time of writing) at \texttt{https://github.com/inouye-lab/explaining-distribution-shifts}.
Both begin by finding a one-to-one optimal transport (OT) mapping between each data point from $P$ (denoted by $x_p$) and a corresponding point from $Q$ (denoted by $x_q$) using the Python Optimal Transport (\texttt{pot}) library.
This requires the datasets for $P$ and $Q$ to be of the same size, so we subsample the larger of the two datasets to make this condition hold.
We also cap the dataset lengths at $20000$ for computational reasons.
Following \citet{kulinski2023explaining}, we one-hot encode categorical features and use the squared Euclidean cost in the OT solver.
Specific details for OT-C and OT-S are given below.

\paragraph{OT-C}

After finding the optimal transport mapping, this baseline groups paired points $(x_p, x_q)$ into $k$ clusters using the standard \texttt{KMeans} implementation from \texttt{scikit-learn}.
Clustering is done in a joint space including feature values from both $x_p$ and $x_q$, which is standardised to have zero mean and unit variance along each feature.
Our addition for this baseline is to compute an attribution for each cluster with respect to the prediction shift of a target model $f$ as follows:
\[
    \phi_{\text{cluster}} \coloneqq \frac{|\text{cluster}|}{N}\mathbb{E}_{(x_p, x_q)\in \text{cluster}}[f(x_q) - f(x_p)],
\]
where $N$ is the total size of each dataset.
By definition, the sum of all cluster attributions is guaranteed to equal the overall prediction shift $\mu_Q - \mu_P$, so there is no meaningful notion of $\text{PercentUnexplained}$ for this baseline.


\paragraph{OT-S}

After finding the optimal transport mapping, this baseline selects the $k$ features with the greatest mean squared difference over all paired points $(x_p, x_q)$.
This is defined as $\mathbb{E}_{(x_p, x_q)} [(x_{p, i} - x_{q, i})^2]$ for each feature $i$.
Our addition is to use the selected feature set, as well as the OT mapping itself, as the basis for an SV analysis.
Let $F$ denote the selected feature set.
Our SV analysis involves performing all subsets of interventions $S\subseteq (F \cup \{\text{OtherFeatures}\})$, where each intervention swaps either one feature from $F$, or \textit{all} features not in $F$, from the value(s) in each $x_p$ to the value(s) in its paired point $x_q$.
We then calculate the mean model prediction over this set of interventional points.
This can be formally defined as follows:
\[
    \hat{\mu}_S \coloneqq \mathbb{E}_{(x_p, x_q)}[f(\text{intervene}(x_p, x_q, S))],
\]
\[
\text{where}\ \ \ \text{intervene}(x_p, x_q, S)_i \coloneqq \left\{\begin{array}{ll}
    x_{q,i} & \text{if }i \in F \text{ and } i\in S, \\
    x_{q,i} & \text{if }i \notin F \text{ and } \text{OtherFeatures}\in S,\\
    x_{p,i} & \text{otherwise.}
    \end{array}\right.
\]
In turn, we compute an SV $\phi_i$ for each $i\in (F \cup \{\text{OtherFeatures}\})$ via the standard formula:
\[
    \phi_{i} \coloneqq \sum_{S\subseteq (F \cup \{\text{OtherFeatures}\} \setminus i)}\frac{|S|!((|F|+1)-|S|-1)!}{(|F|+1)!}\Big[\hat{\mu}_{S\cup\{i\}} - \hat{\mu}_{S}\Big].
\]
Furthermore, we define a $\text{PercentUnexplained}$ value in an analogous way to our own method:
\[
    \text{PercentUnexplained}_{\text{OT-S}} \coloneqq 100 \times |\phi_{\text{OtherFeatures}}| / |\mu_Q - \mu_P|.
\]
This value captures the proportion of the prediction shift that cannot be accounted for by shifts along the selected features $F$.

\section{Additional Experiments}

\subsection{$\text{PercentUnexplained}$ Results for Gradient Boosted Tree Ensemble on \textit{Breast Cancer Wisconsin}}
\label{app:extra_gb}

\begin{figure}[h!]
    \centering
    \includegraphics[width=0.95\textwidth]{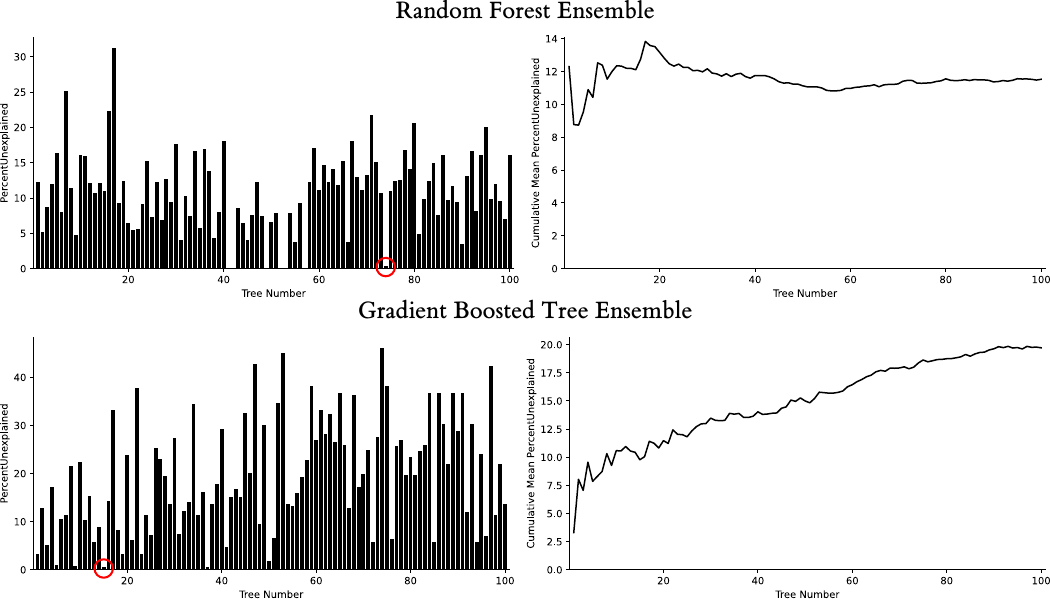}
    \caption{$\text{PercentUnexplained}$ results for random forest and gradient boosted tree ensembles.}
    \label{fig:gb_vs_rf}
\end{figure}

In the top-left subplot of Figure~\ref{fig:gb_vs_rf}, we reprint the $\text{PercentUnexplained}$ results for all $100$ $8$-leaf trees in a random forest ensemble for the \textit{Breast Cancer Wisconsin} dataset.
Below that, we show equivalent results for a gradient boosted tree ensemble (specifically an \texttt{xgboost} \texttt{XGBClassifier}), also comprised of $100$ $8$-leaf trees.
The minimum $\text{PercentUnexplained}$ values across the random forest and gradient boosted ensembles are very similar, at $0.4\%$ and $0.5\%$ respectively.
The means differ more widely, at $11.5\%$ and $19.7\%$ respectively.

The most interesting point of contrast, however, is the sequential trend across the tree numbers in the ensemble.
While values vary significantly from each tree to the next, earlier trees in the sequence tend to have lower $\text{PercentUnexplained}$ values for the gradient boosted model.
No such trend exists for the random forest.
This difference is shown more clearly by plotting the cumulative mean $\text{PercentUnexplained}$ over the tree sequence, shown in the right-hand subplots of Figure~\ref{fig:gb_vs_rf}.

We hypothesise that this trend is due to how gradient boosted ensembles are constructed.
The first tree in the sequence (tree~$1$) is trained to directly solve the prediction task, tree $2$ is trained to compensate for errors in the first, tree $3$ to compensate for errors on the first two, and so on.
Hence, later trees in the sequence are optimised for increasingly esoteric compensation tasks that may focus on increasingly small subsets of the training dataset.
In turn, this usually (but not always) means the splits of those trees define less useful subgroups for explaining ensemble-wide prediction shift at the level of entire distributions.

Regardless of the precise causality, the trend of increasing $\text{PercentUnexplained}$ over gradient boosted tree sequences is one that we observe consistently.
It creates an opportunity to implement an early-stopping mechanism, whereby we only perform the SV analysis on the first few (e.g. $20$) trees to lower the computational expense.
For the ensemble in Figure~\ref{fig:gb_vs_rf}, that early-stopping threshold would be sufficient to identify the tree with the single lowest $\text{PercentUnexplained}$ value, which is tree $14$.


\section{Limitations of Method} \label{app:limitations}

Our method inherits the theoretical advantages of Shapley values via the four SV axioms (efficiency, symmetry, linearity and null player), but it also inherits its limitations, most importantly the exponential complexity with the number of factors. This means that applying the exact method to large trees is impractical. However, our results show that (1) Kernel SHAP and pruning approximations can be used effectively to greatly reduce the complexity when large trees are given, and (2) small surrogate trees can be sufficient to explain prediction shifts in neural networks with low PercentUnexplained. A separate limitation of SVs is the subtlety of their interpretation. The attribution given to any given factor in an SV analysis depends on the set of other factors under consideration, rather than being intrinsic to that factor only. This means that a fully honest presentation of our results should always present all factors.

While the use of conditional probabilities is crucial to enable a well-posed SV calculation (see Appendix~\ref{app:why_subgroup}), it does create a possibility for certain probabilities to be undefined on distribution $P$, $Q$ or both, preventing our method from working. As discussed in Appendix \ref{app:failure_mode}, this occurs very rarely in practice for the tree-specific variants of our method, and direct control of the surrogate growth process means we can ensure it \textit{never} occurs for the model-agnostic variant.

While our empirical results indicate that our method can usually achieve low $\text{PercentUnexplained}$ values, there is no guarantee that this will be the case unless the target model is a single decision tree (in which case $\text{PercentUnexplained}$ is always zero). This means that there may be certain domains, models and shifts for which our method can provide little useful insight. Crucially, however, the fact that the $\text{PercentUnexplained}$ is an explicit output of our SV analysis means that \textit{we will always know when this is the case} and can disregard the results accordingly. The fact that our method is able to explicitly estimate its own limitations is, from this perspective, a marked advantage.

For tree-based models, we know that the conditional factors used in our method correspond to true causal mechanisms that drive model predictions. In the model-agnostic case, however, the use of an external surrogate means that such a direct causal link is not guaranteed. Certain conditional factors could be deemed important by our SV analysis despite the associated input features not actually being used by the target model; this could result from statistical correlations in the observed data from $P$ and $Q$. That said, identifying such cases of unfaithful explanations is precisely the aim of our faithfulness evaluations (R-Faithfulness and activation curve analysis). Figure~\ref{fig:baseline_comparison_quant} indicates that our model-agnostic method achieves superior faithfulness results compared to the OT baselines.

Separately, even in the tree-specific case where we know which conditionals are causal of model behaviour, further domain-specific analysis is needed to understand why the conditionals themselves have changed. For example, in Figure~\ref{fig:ensemble_experiment}, we attribute a reduction in predicted breast cancer malignancy to an increase in cases with $\leq 2$ normal nucleoli. Why has this increase occurred? Is there an important biological difference, or some change in the testing regime? Such follow-up questions are completely valid and crucial for clinical decision-making, but in an important sense, they are beyond the scope of our method. Answering questions about \textit{why} observed changes in data occur would require a causal model of the underlying domain, to which we do not assume access in our work. One should think of methods like ours as providing an an explanatory starting point, informing which questions to ask when applying domain knowledge to understand the causal origins of prediction shifts. Integrating our method with causal domain knowledge is an exciting direction for future work.

Finally, the current implementation of our method requires a finite-dimensional input space of individually-interpretable numerical or categorical features, over which decision trees can be learnt. This prevents its immediate application to other data modalities such as text, images and audio.
As discussed in the conclusion of the main paper, it would be extremely interesting to consider how this gap could be bridged via feature extraction and representation learning techniques.


\newpage

\end{document}